\documentclass[11pt]{article}
\usepackage{coling2018}
\usepackage{times}
\usepackage{url}
\usepackage{latexsym}
\usepackage{graphicx}
\usepackage{caption}
\usepackage{subcaption}

\title{Principled Frameworks for Evaluating Ethics in NLP Systems}

\author{
Shrimai Prabhumoye, Elijah Mayfield, Alan W Black \\
  Language Technologies Institute, Carnegie Mellon University \\
  Pittsburgh, PA, USA \\
  {\tt sprabhum, emayfiel, awb@andrew.cmu.edu} 
  }

\date{}

\begin{document}
\maketitle
\begin{abstract}
We critique recent work on ethics in natural language processing. Those discussions have focused on data collection, experimental design, and interventions in modeling. But we argue that we ought to first understand the \textit{frameworks of ethics} that are being used to evaluate the fairness and justice of algorithmic systems. Here, we begin that discussion by outlining deontological ethics, and envision a research agenda prioritized by it.
\end{abstract}

Due to the sheer global reach of machine learning and NLP applications, they are empowered to impact societies \cite{hovy16} - potentially for the worse. Potential harms include exclusion of communities due to demographic bias, overgeneralization of model predictions to amplify bias or prejudice, and overstepping privacy concerns in the pursuit of data and quantification \cite{mieskes2017quantitative}. 

Many researchers are trying to make sense of these topics. \newcite{crawford2017trouble} give us theory to work from, presenting \textit{allocational harm} and \textit{representational harm}; \newcite{lewis2017integrating} examines the role of government regulation on accountability in ethics; \newcite{smiley2017say} opens a discussion on ethics checklists for acceptance testing and deployment of trained models. All of these works identify potential ethical issues for NLP, and all propose best practices for data collection and research conduct. But presently there is no external accountability for which approach to ethical NLP is correct - as machine learning researchers, we are evaluating ourselves, on metrics of our own choosing. Much of the existing work on ethics in NLP is \textit{normative} - evaluation of whether a system ``does the right thing." We argue that before the field can establish normative goals, we need to reason about \textit{meta-normative} decisions: specifically, how do we even decide what it means to be ``right"?

We believe that it is time for a more impartial arbitration of ethics in our field, emphasizing the need for a grounding in  frameworks that long predate the questions we're faced with today. By reaching out to other fields, we keep the question of ethics at arms length from our own work, giving us a neutral playing field on which to judge ethical performance of machine learning systems. Philosophy has much to offer us; we describe two competing frameworks: the \textit{generalization principle} and the \textit{utilitarian principle}.

\paragraph{Ethics under the generalization principle:}
\begin{quote}
\textit{[An ethical decision-maker] must be rational in believing that the reasons for action are consistent with the assumption that \textbf{everyone} with the same reasons will take the same action.}\footnote{From \newcite{hooker2018truly}.} 
\end{quote}

This approach is founded on the work of \newcite{kant1785}, which fundamentally prioritizes \textit{intent} as the source of ethical action. 
To analyze this in machine learning, we state that a trained agent $A$ is expected to take an action $d_i$ based on a given set of evidence $E_i$, from a finite closed set of options $D$. 
This simple notation can be extended to classification, regression, or reinforcement learning tasks. 
The generalization principle states that agent $A$ is ethical if and only if, when given two identical sets of evidence $E_1$ and $E_2$ with the \textit{same} inputs, agent $A$ chooses to make same decision $d_1$ every time. Furthermore, the principle assumes that all \textit{other} such trained agents will \textit{also} make those same predictions.

Here, we presume that the input representation is sufficient to 
make a prediction, without including any extraneous information.
The reasons for an act define the \textit{scope} of the act, or the set of necessary and sufficient conditions under which that act is generalizably moral \cite{hookerEthics}. 
Evidence must be relevant to the decision making process, and moreso must exclude task-irrelevant evidence that might be a source of bias. 
By excluding such evidence, our agent is invariant to \textit{who} is being evaluated, and instead focuses its decision solely on task-relevant evidence.

This goal cannot be met without transparent and sparsely weighted inputs that do not use more information than is necessary and task-relevant for making predictions. Practically, this definition would privilege research on interpretable, generalizable, and understandable machine learning classifiers. The burden of proof of ethics in such a framework would lie on transparency and expressiveness of inputs, and well-defined, expected behavior from architectures for processing those features. Some work on this - like that from \newcite{hooker2018toward} - has already begun. If deontological ethics were prioritized, we would expect to see rapid improvement in parity of $F_1$ scores across subgroups present in our training data - an outcome targeted by practitioners like \newcite{chouldechova2017fair} and \newcite{corbett2017algorithmic}.

\paragraph{Ethics under the utilitarian principle:}
\begin{quote}
\textit{An action is ethical only if it is not irrational for the agent to believe that no other action results in greater expected utility.} \footnote{From \newcite{hookerEthics}.}
\end{quote}

In this formulation, which can be traced back to \newcite{bentham1789}, 
an algorithmic system is expected to understand the consequences of its actions.
We measure systems by whether they maximize total overall welfare in their \textit{results}. 
We once again train an agent $A$, which will make a decision $d_i$ for each evidence set $E_i$. 
But here, we also assign a utility penalty or gain $u_i$ for each of those decisions. 
Rather than judge the algorithm based on whether it followed consistent rules, we instead seek to maximize \textit{overall} gain for all $N$ decisions that would be made by agent $A$ - morality of an agent is equal to $\Sigma_i^N u_i $.

This is a very different worldview! 
Here, the burden of provable ethical decision-making no longer lands on transparency in the algorithm or consistency of a classifier over time. 
Instead, proof of ethical behavior rests on our ability to observe the consequences of the actions the agent takes. 
One could argue that consequences are hard to estimate and hence we can pick a random action.
But that would be irrational.
Hence, the principle judges an action by whether the agent acts according to its rational belief of maximizing the expected utility, rather than by the actual consequences.
If the agent is wrong then the action turns out to be a poor choice, but nonetheless ethical because it was a rational choice.

In \newcite{crawford2017trouble}, the author appeals to researchers to actively consider the subgroups that will be harmed or benefited by the automated systems. 
This plotting of expected consequences and their exhaustive measurement takes precedence in utilitarian ethics, de-prioritizing the interpretability or transparency of the learned model or features that govern our agent. 
For machine learning researchers, this would mean shifting the focus toward building rich and exhaustive test datasets, cross-validation protocols, and evaluation suites that mirror real-world applications to get a better measurement of impact. 

From this work, we might see an initial drop in reported accuracy of our systems as we develop broader test sets that measure the utility of our systems; however, we would then expect overall accuracy on those broad test sets to be the primary measure of ethical fitness of the classifiers themselves. Subgroup-based parity metrics would fall by the wayside in favor of overall accuracy on data that mirrors the real world.

\paragraph{Real World Scenarios:}

\begin{figure}
\centering
\begin{subfigure}{.5\textwidth}
  \centering
  \includegraphics[width=0.8\linewidth]{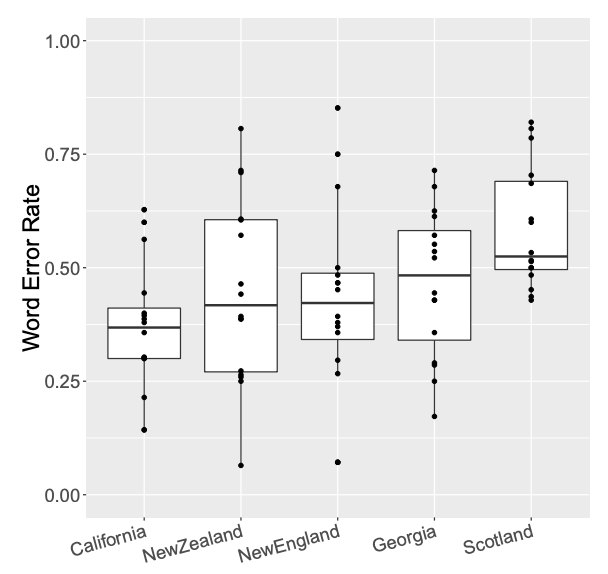}
  \caption{YouTube automatic caption word error
rate by \\ speaker’s dialect region. Points indicate
individual speakers.}
  \label{fig:sub1}
\end{subfigure}%
\begin{subfigure}{.5\textwidth}
  \centering
  \includegraphics[width=.75\linewidth]{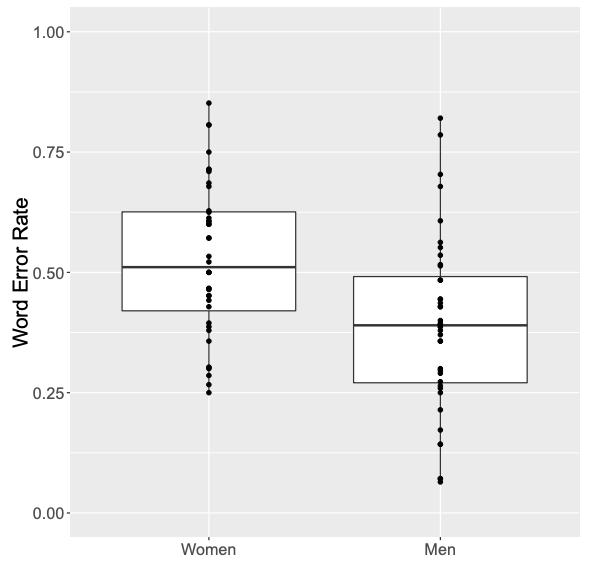}
  \\
  \caption{ YouTube automatic caption word error
rate by speaker’s \\ gender. Points indicate individual speakers.}
  \label{fig:sub2}
\end{subfigure}
\caption{Word Error rate plots for gender and dialect \cite{tatman2017gender}}
\label{fig:asr}
\end{figure}

These philosophical frameworks do not always diverge in their evaluation of models. Sometimes, models have unambiguously unethical gaps in performance. The exploration from \newcite{tatman2017gender}, for instance, shows the difference in accuracy of YouTube's automatic captioning system across both gender and dialect with lower accuracy for women and speakers from Scotland (shown in Figure \ref{fig:asr}, reproduced from the original work). This study shows how this system violates the utilitarian principle by negatively impacting the utility of automatic speech recognition for women and speakers from Scotland. YouTube's model also violates the generalization principle, by incorporating superfluous information about speakers in the representation space of the models. The authors suggest paths forward for improving those models and show that there is room to improve.

But sometimes, solutions highlight differences across ethical frameworks. In \newcite{hovy2015demographic}, for instance, the author shows that text classification tasks, both sentiment and topic classification, benefit from embeddings that include demographic information (age and gender). Here, the two ethical frameworks we have discussed diverge in their analysis. The generalization principle would reject this approach: age and gender shouldn't intrinsically be used as part of a demographic-agnostic topic classification task, if the number of sources of information is to be minimized. Similarly, changing the feature space depending on the author, rather than the content of the author's text, does not result in models that will make the same decision about a text independent of the identity of the author. The utilitarian principle, in contrast, aligns with the Hovy approach. A more accurate system benefits more people; incorporating information about authors improves accuracy, and so including that information at training and prediction time increases the expected utility of the model, even if different authors may receive different predictions when submitting identical texts.

For an alternate example in which the generalization principle was prioritized over utility, consider the widely-cited coreference resolution system of \newcite{bolukbasi2016man}. This paper found that word embeddings used for coreference resolution were incorporating extraneous information about gender - for instance, that doctors were more likely to be men, while nurses were more likely to be women. This and similar work in ``debiasing" word embeddings follows the generalization principle, arguing that removing information from the embedding space is ethically the correct action, even at the expense of model accuracy. The authors of the Bolukbasi work do find that they can minimize the drop in expected utility, reducing F1 scores by less than 2.0 while removing stereotypes from their model. However, in a fully utilitarian ethical framework, even this drop would be unjustifiable if the model simply reflected the state of the world, and removing information led to reduced performance.


\paragraph{Call to Action:}

As a field, we risk building an incoherent set of research on fairness and ethics if we do not address these questions early. We recommend researchers ground their work in philosophical theory, rather than in arbitrary measurement of metrics invented for and by ourselves. We recommend reuse and replication of these methods, to ensure a common vocabulary and language within our field. And while we do not take a stance on the sole correct ethical framework to follow - a debate going back at least to Aristotle - we targue for a rational discourse that acknowledges history and leads our field to a richer definition of fairness and ethics, for the sake of better systems that we put out into the world. 





\bibliography{winlp}
\bibliographystyle{acl}




\end{document}